\documentclass{isprs} % isprs class modified 23-04-2019 (Dennis Wittich)
\usepackage{hyperref}
\usepackage{subfigure}
\usepackage{floatrow}
\usepackage{setspace}
\usepackage{geometry} % added 27-02-2014 Markus Englich
\usepackage{epstopdf}
\usepackage{bbding} %\Checkmark
\usepackage[labelsep=period]{caption}  % added 14-04-2016 Markus Englich - Recommendation by Sebastian Brocks
\usepackage[british]{babel} 
\usepackage[hang]{footmisc}
\usepackage[round, authoryear]{natbib}

\makeatletter
\newcommand{\bianca}{\renewcommand\NAT@open{[}\renewcommand\NAT@close{]}}
\makeatother

\begin{document}

\title{Towards Large-scale Building Attribute Mapping using Crowdsourced Images: Scene Text Recognition on Flickr and Problems to be Solved
}

% KAO: Remove extra spacing
\author{
 Y. Sun\textsuperscript{1}\thanks{Corresponding author} ,  
 A. Kruspe\textsuperscript{2}, 
 L. Meng\textsuperscript{3}, 
 Y. Tian\textsuperscript{1},
 E. J. Hoffmann\textsuperscript{1},
 S. Auer\textsuperscript{4}, 
 X. X. Zhu\textsuperscript{1}}

% KAO: Remove extra newline
\address{
\textsuperscript{1} Data Science in Earth Observation, Technical University of Munich, \\(yao.sun, yifan.tian, eike.jens.hoffmann, xiaoxiang.zhu)@tum.de\\ 
\textsuperscript{2 } Falculty of Computer Science, Technische Hochschule Nürnberg, anna.kruspe@th-nuernberg.de\\
\textsuperscript{3 } Cartography and Visual Analytics, Technical University of Munich, liqiu.meng@tum.de\\
\textsuperscript{4 } Remote Sensing Technology Institute, German Aerospace Center (DLR), stefan.auer@dlr.de\\
}

% If the corresponding author is NOT the final author, always add a % space before the subsequent comma, i.e.
% first author name\textsuperscript{a,}\thanks{Corresponding author} , % second author name \textsuperscript{b}, etc.
% thanks to Niclas Borlin 05-05-2016

% the use of the information of commissions and working groups should not be used any longer and has been commented out
% C. Heipke, Sept. 20,2022
%\commission{XX, }{YY} %This field is optional. If filled, XX and YY should be replaced by adequate numbers. See https://www2.isprs.org/commissions/
%\workinggroup{XX/YY} %This field is optional.
%\icwg{}   %This field is optional.

\abstract{
Crowdsourced platforms provide huge amounts of street-view images that contain valuable building information. This work addresses the challenges in applying Scene Text Recognition (STR) in crowdsourced street-view images for building attribute mapping. We use Flickr images, particularly examining texts on building facades. A Berlin Flickr dataset is created, and pre-trained STR models are used for text detection and recognition. Manual checking on a subset of STR-recognized images demonstrates high accuracy. 
We examined the correlation between STR results and building functions, and analysed instances where texts were recognized on residential buildings but not on commercial ones. 
Further investigation revealed significant challenges associated with this task, including small text regions in street-view images, the absence of ground truth labels, and mismatches in buildings in Flickr images and building footprints in OpenStreetMap (OSM). 
To develop city-wide mapping beyond urban hotspot locations, 
we suggest differentiating the scenarios where STR proves effective while developing appropriate algorithms or bringing in additional data for handling other cases. 
Furthermore, interdisciplinary collaboration should be undertaken to understand the motivation behind building photography and labeling. The STR-on-Flickr results are publicly available at \url{https://github.com/ya0-sun/STR-Berlin}.

}

\maketitle
%\saythanks % added 28-02-2014 Markus Englich

\section{Introduction}\label{MANUSCRIPT}
 
% KAO: Sloppy spacing ensures non-overfull lines. Can be removed if this is not an issue.
\sloppy

Building information extraction has been a hot topic since three decades in photogrammetry and remote sensing. Numerous studies have been conducted using different types of data, e.g., optical satellite images~\citep{liasis2016satellite}, airborne LiDAR~\citep{brenner2005building}, Synthetic Aperture Radar~\citep{sportouche2011Extraction, brunner2010Building, sun2021bbox}. 
However, most of the research efforts primarily concentrate on building geometries, such as footprints, heights, and 3D models in different levels of details~\citep{li2020instance, sun2020cgnet, chen2021maskH, sun20163d}, and hardly attend to building attributes, such as building type, age, material, ownership, number of households, and more, which are essential for varies urban application such as public facility planning and resource distribution.

Some works employ aerial or satellite images to estimate building attributes, e.g., building functions~\citep{huang2018urban, zhang2019joint}. However, nadir-looking images are inherently ambiguous as they mainly feature rooftops. 
In recent years, researchers started to employ street-view images (SVI) featuring building facades. 
Google Street View images, as a major commercial source, have been used for building age estimation~\citep{li2018estimating}, flood risk of buildings~\citep{chen2022deep}, building heights~\citep{yan2022estimation}, and more. 
On the other hand, 
crowdsourced platforms, such as Flickr, Unsplash, and Mapillary, provide huge amounts of street-view images containing valuable information.  
They are ubiquitous, cheap, easy to collect, and increasingly prevalent in research. 
Flickr, as an example, has been employed for mapping and understanding landscape aesthetics~\citep{langemeyer2018mapping}, land use classification~\citep{leung2012exploring}, and flood-level estimation~\citep{chaudhary2019flood}. 
As for building attributes, Kang et al. \citep{kang2018building} classified building instances from street-view images using convolutional neural networks. 
Hoffmann et al. \citep{hoffmann2023using} employed Flickr images from 42 cities to classify buildings using deep neural networks, demonstrating the mapping potential utilizing crowdsourced data on a large scale. 

However, in image classification tasks, texts are often ignored. Texts on building facades contain rich attribute information, such as shop names, building usage, house numbers, and construction years. 
Scene Text Recognition (STR) is the task of reading texts in natural scenes. Despite the success of Optical Character Recognition (OCR) systems on clean documents, the STR remains a difficult task due to the diverse text appearances in the real world and the imperfect conditions in which these scenes are captured. 
Sun et al. \citep{sun2023ai} presented STR results and extracted building attributes from a few street-view images, however, large-scale mapping remains challenging. 

Aiming at large-scale building attribute mapping using crowdsourced images, we apply STR on Flickr data. 
In this paper, 
we report our preliminary results and observations to identify situations in which STR-on-SVI helps map building attributes and address the challenges in this field for future works. 

Next, we present the methods in Section~\ref{metho} 
and detail the experimental results and analysis in Section~\ref{exd}. Finally, Section~\ref{co} concludes the paper and outlines possible future directions.

\section{Methodology}\label{metho}
Our workflow comprises three main steps: first, build the Flickr dataset; second, associate building functions with the Flickr images; and third, extract texts on the images using STR.

\subsection{Flickr image filtering}

Flickr covers a diverse range of content and motifs and provides an accessible API encouraging users to share and use photos.  
For our STR tasks, it is necessary to filter out images unrelated to buildings and images without a valid geotag or compass orientation. Therefore, a filtering pipeline is designed to identify images in the Flickr dataset that meet these criteria. For more details, interested reader is referred to~\citep{hoffmann2023using}. Next, we briefly explain it.

\subsubsection{Content filtering}

This step filters out images containing no buildings. It comprises Google Street View similarity filtering and object detection filtering. 

First, we filter out non-street-view images from the Flickr dataset. The problem is approached as an image retrieval task, utilizing Google Street View images as the seed dataset and a Flickr dataset. Deep neural network features are utilized for identifying structurally similar images, extracting features from the last hidden layer of a pre-trained VGG16 network on  ImageNet~\citep{russakovsky2015imagenet}. Cosine similarity is calculated on the resulting feature vectors. 
For the seed dataset, pre-calculated features are used. 
Then, pairwise cosine similarity is determined between Flickr images and the seed dataset. Images with similarity parameters below a predefined threshold are discarded. 

Next, we apply object detection algorithm to ensure the presence of building facades in Flickr images. The algorithm identifies objects in the previously filtered images, generating a list of objects for each image. If this list contains a house or a building with a size parameter greater than the threshold and a confidence score higher than the threshold, the image qualifies for further processing.

\subsubsection{Metadata filtering}
This step filters out images that cannot be geo-located. 
It focuses on the image's position and compass direction, that are crucial for calculating a line of sight for matching Flickr images with building footprints. 

Geotags can be created automatically by a GPS sensor in the camera or manually by the user, and the latter is often inaccurate since users tend to tag images batch-wise, leading to slight position differences between GPS-tagged images taken without significant movement. To identify images with manually added geotags, we employ a heuristic filter. If two images have the same position, manual tagging without GPS data is suggested. These images are omitted from the subsequent steps. 

Next, we check the metadata on the standard EXIF\footnote{EXIF is a standard established by the Camera and Imaging Products Association (CIPA) and the Japan Electronics and Information Technology Industries Association (JEITA).}, which includes details like the capture date, camera model, settings, and GPS sensor data, such as latitude, longitude, and compass direction. 
The presence of the GPSImgDirection tag in the EXIF data is checked, and images lacking this tag are rejected.

\subsection{Mapping OpenStreetMap (OSM) building functions to Flickr images}

\subsubsection{Building function aggregation in OSM }

We obtain building footprints and their tags in OSM in the study area. 
OSM allows users to contribute mapping data in a Wikipedia-like manner. Though OSM provides guidelines for structuring and enriching data, there is no strict enforcement. Consequently, building tags are optional, with only building footprint coordinates being mandatory when added to OSM. OSM guidelines include three tags: building, amenity, and shop, used to indicate building functions. 

We implement a classification scheme based on OSM guidelines, assigning each tag value (building, amenity, and shop) to one of three categories: commercial, residential, or other. If multiple tags are present, we ensure they are consistent in their classification. Disagreements among tags result in the building being unmapped to any class. However, if only one tag or all available tags agree on the same class, we assign that class to the building.

\subsubsection{Matching Flickr images and OSM buildings}

In this step, we connect the buildings depicted in an image and their corresponding building footprints in OSM.

We utilize the image's position and compass direction from the EXIF data, which is essential for creating a line of sight. The line of sight identifies possible building candidates by intersecting with their polygons in OSM. From these candidates, we select the building with the closest distance to the image's position as the reference building. 

\subsection{Information extraction on Flickr images}

\subsubsection{STR on Flickr}
STR algorithms are applied on the Flickr images to extract texts on buildings, specifically, TextSnake~\citep{long2018textsnake} for text detection and SAR~\citep{li2019show} for text recognition:

\begin{enumerate}
    \item{Text detection: }
    TextSnake~\citep{long2018textsnake} is a novel approach for text detection in natural scenes. Unlike conventional methods that represent text regions as bounding boxes or polygons, TextSnake models text instances as a sequence of pixels forming snakes. It combines convolutional neural networks and recurrent neural networks to predict text instances' positions and orientations simultaneously. The TextSnake method performs better in handling curved and arbitrarily shaped text instances, making it highly effective for scene text detection tasks. 
    
    \item{Text recognition: }
    SAR~\citep{li2019show} is based on an attention-based encoder-decoder framework. The model leverages visual attention mechanisms to focus on relevant regions of the input image, allowing it to recognize irregularly shaped and oriented text instances adaptively. The system demonstrates exceptional performance across various challenging scenarios, such as curved and distorted text, making it a robust and efficient baseline for addressing irregular text recognition tasks.
\end{enumerate}

\subsubsection{STR results filtering}
As no ground truth labels are available for the texts in the Flickr images, we apply the following criteria to filter the results obtained from the STR process. The aim is to eliminate STR results with lower confidence levels.

\begin{enumerate}
    \item{Text score and box score:}
    Each recognized text in STR is associated with a box score and a text score, indicating the confidence of the detection and the recognition, respectively. Both the scores range from 0 to 1, and the larger number indicates higher confidence. Therefore, we filter the results using pre-defined thresholds on both scores.     
    
    \item{Stopwords:} 
    Stopwords are common words that do not carry significant meaning, such as “the,” “is,” and “and,”. While essential for sentence structure, they can add noise during text analysis. Filtering them out helps focus on meaningful words, improve efficiency, enhance relevance, and boost search accuracy. 
    Since stopwords are unlikely frequently appear on building facades, we filter them out in STR results as misrecognition.

    \item{Repetitive letters:}
    Lastly, we filter out text strings containing repetitive letters that are not words and are misrecognized from building structures, such as windows and balcony railings. 
    
\end{enumerate}

\section{Experiments and Discussion}\label{exd}

\subsection{Data and study area}
The Flickr dataset in Berlin was used in our experiments, containing 3,431 street-view building images filtered from 929,508 Flickr images queried using the Flickr API. 
After matching the images with OSM building footprints, 1,833 (53.42\%) are labeled as residential, 605 (17.63\%) as commercial, and 993 (28.94\%) as other.

\begin{figure*}[t]
    \centering    
    \subfigure{\includegraphics[width=0.24\linewidth]{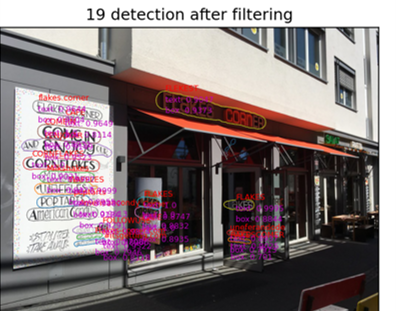}}\hfill
    \subfigure{\includegraphics[width=0.24\linewidth]{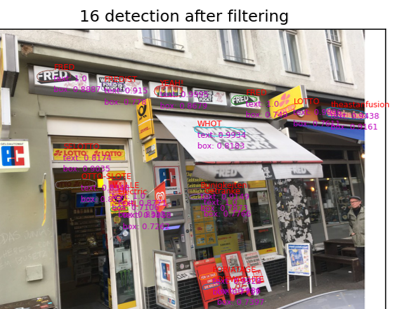}}\hfill
    \subfigure{\includegraphics[width=0.24\linewidth]{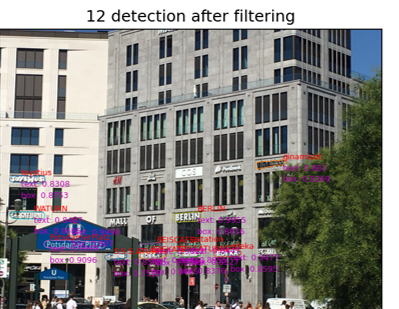}}\hfill
    \subfigure{\includegraphics[width=0.24\linewidth]{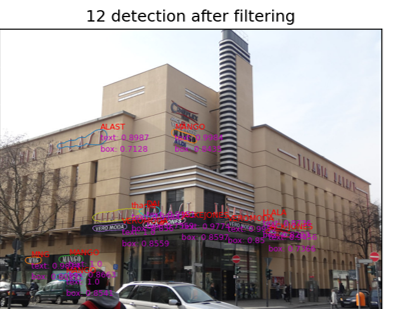}}\hfill
    \\
    \subfigure{\includegraphics[width=0.24\linewidth]{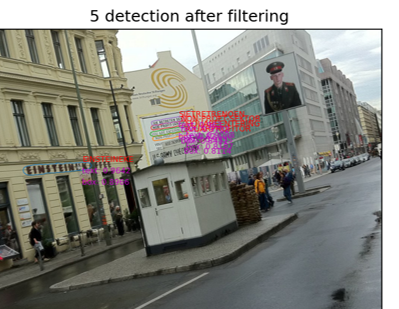}}\hfill
    \subfigure{\includegraphics[width=0.24\linewidth]{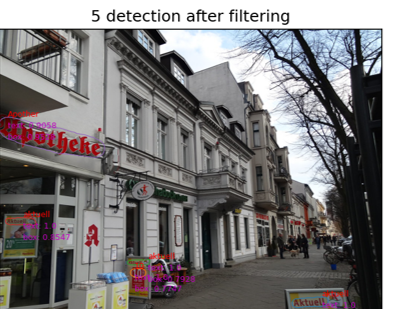}}\hfill
    \subfigure{\includegraphics[width=0.24\linewidth]{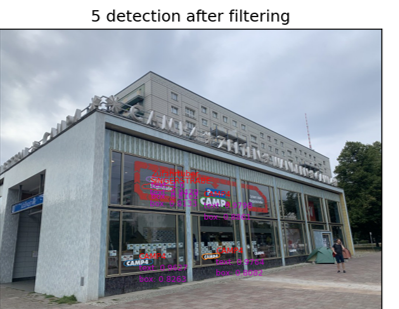}}\hfill
    \subfigure{\includegraphics[width=0.24\linewidth]{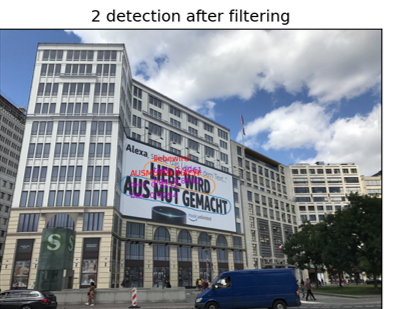}}\hfill
    \\
    \subfigure{\includegraphics[width=0.24\linewidth]{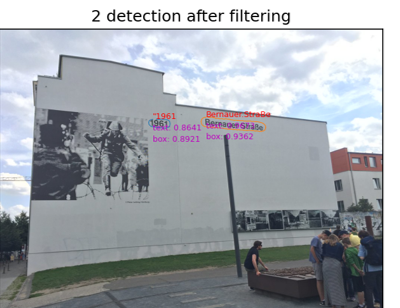}}\hfill
    \subfigure{\includegraphics[width=0.24\linewidth]{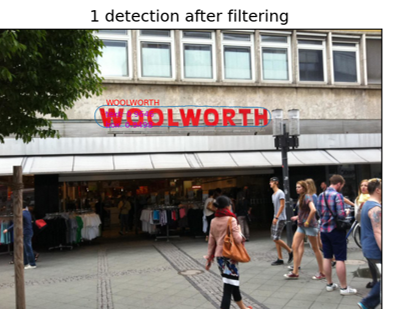}}\hfill
    \subfigure{\includegraphics[width=0.24\linewidth]{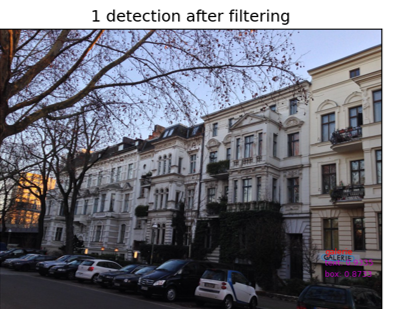}}\hfill
    \subfigure{\includegraphics[width=0.24\linewidth]{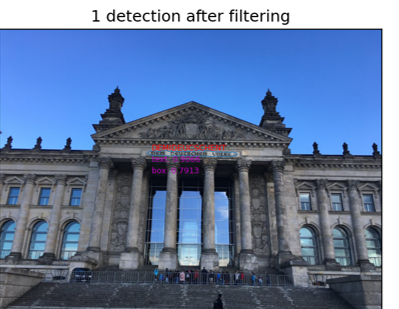}}\hfill
    \\
    \subfigure{\includegraphics[width=0.149\linewidth]{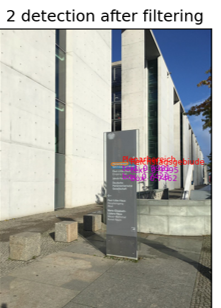}}\hfill
    \subfigure{\includegraphics[width=0.149\linewidth]{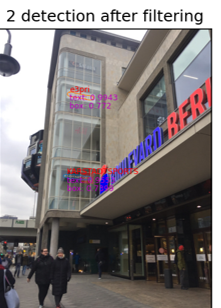}}\hfill
    \subfigure{\includegraphics[width=0.149\linewidth]{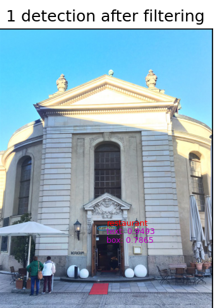}}\hfill
    \subfigure{\includegraphics[width=0.149\linewidth]{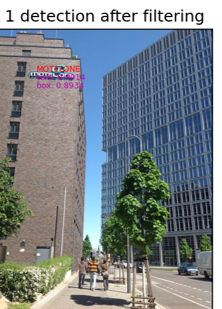}}\hfill
    \subfigure{\includegraphics[width=0.164\linewidth]{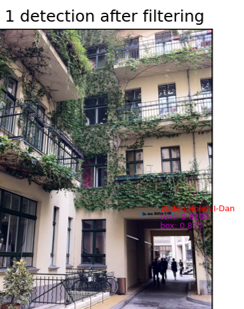}}\hfill
    \subfigure{\includegraphics[width=0.149\linewidth]{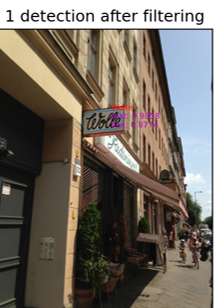}}\hfill
    \caption{Examples of STR results on Flickr building images. The number of detection after filtering is written above each image and the STR results, i.e., texts/text scores and box/box scores, are displayed on the images at the corresponding location. }
    \label{fig:str}
\end{figure*}

% \begin{figure*}[th]
%     \centering    
%     \subfigure[]{\includegraphics[width=0.26\linewidth]{figs/pn-1.png}}\hfill
%     \subfigure[]{\includegraphics[width=0.265\linewidth]{figs/pn-2.png}}\hfill
%     \subfigure[]{\includegraphics[width=0.33\linewidth]{figs/pn-3.png}}
%     \\
%     \subfigure[]{\includegraphics[width=0.33\linewidth]{figs/pn-4.png}}\hfill
%     \subfigure[]{\includegraphics[width=0.33\linewidth]{figs/pn-5.png}}\hfill
%     \subfigure[]{\includegraphics[width=0.33\linewidth]{figs/pn-6.png}}
%     \caption{Examples of Flickr images with only number strings in STR results. }
%     \label{fig:num}
% \end{figure*}

\begin{figure*}[!]
    \centering    
    \subfigure[`19']{\includegraphics[width=0.165\linewidth]{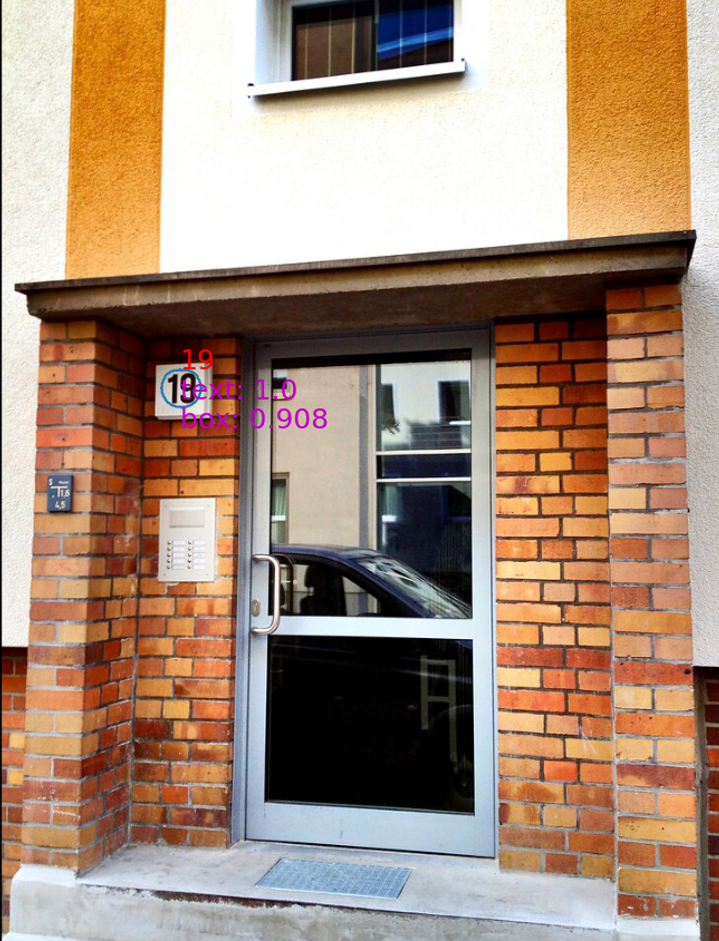}}\hfill
    \subfigure[`1952']{\includegraphics[width=0.17\linewidth]{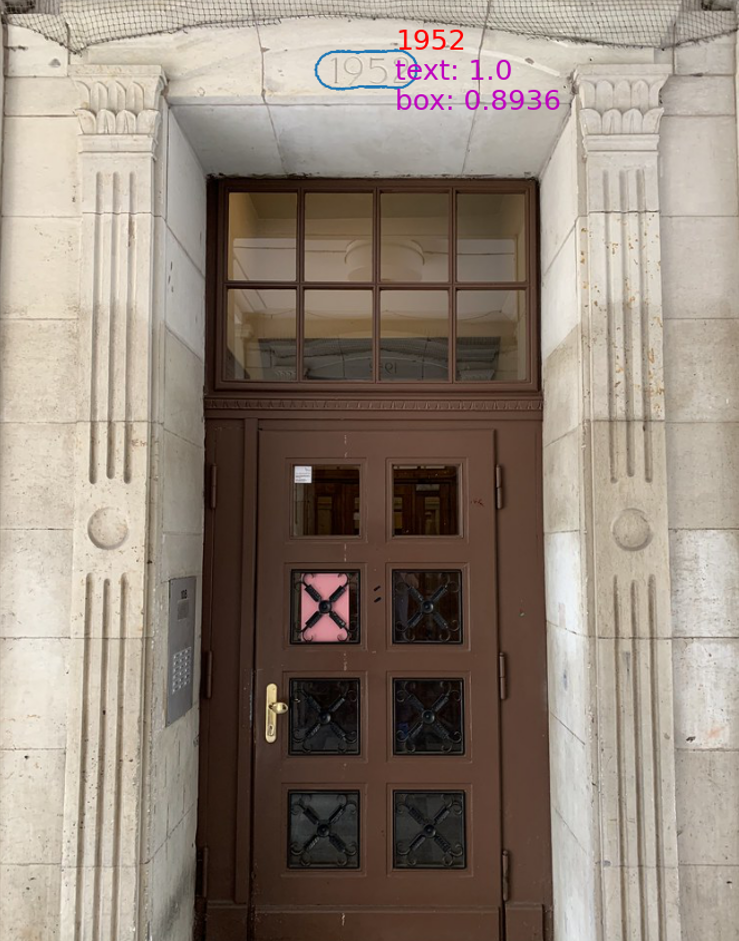}}\hfill
    \subfigure[`1990']{\includegraphics[width=0.21\linewidth]{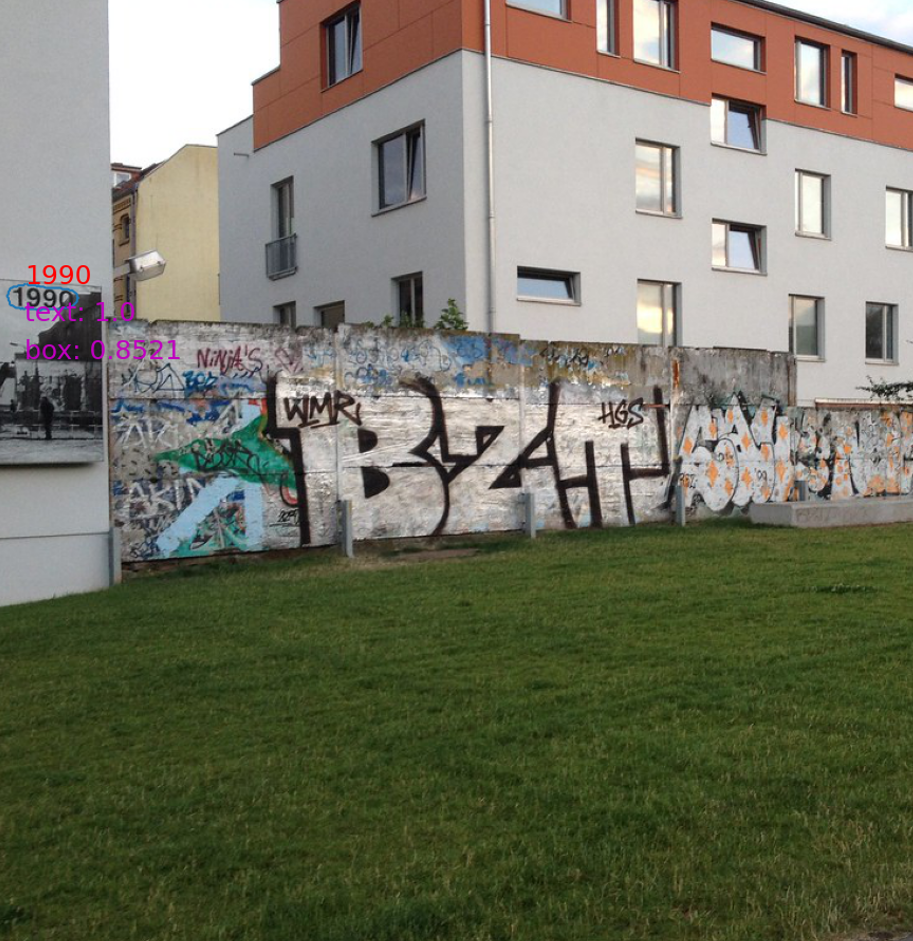}}\hfill
    \subfigure[`30']{\includegraphics[width=0.21\linewidth]{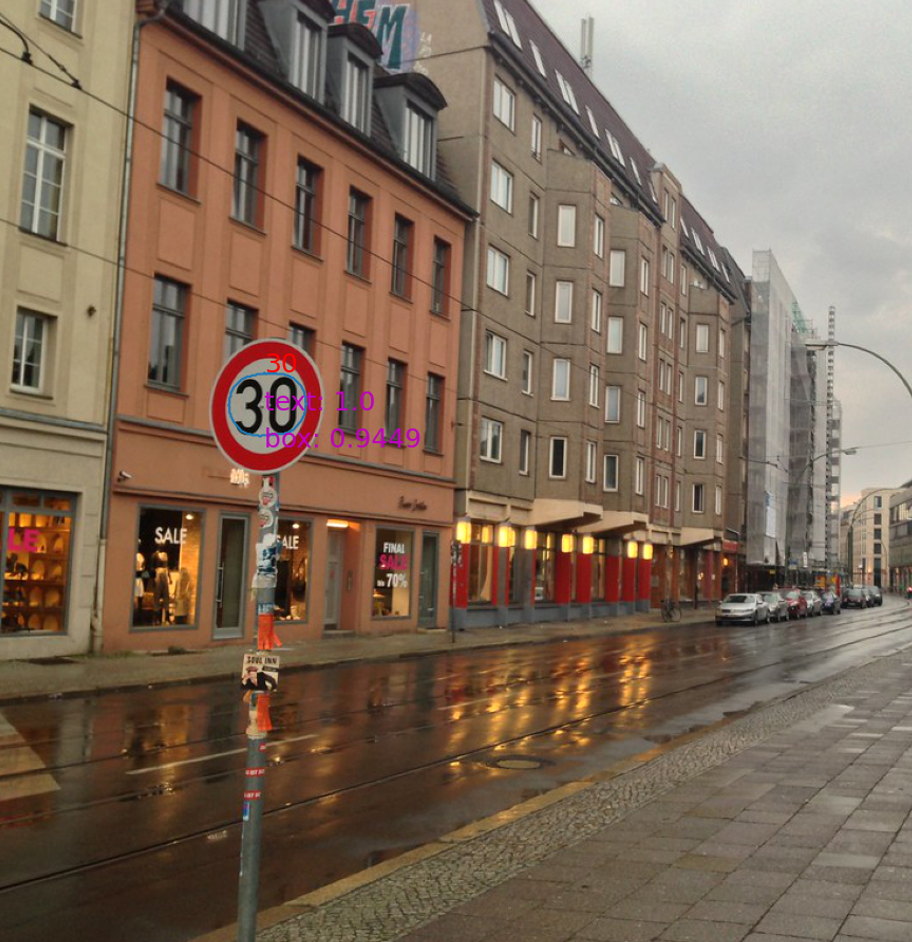}}\hfill
    \subfigure[`10']{\includegraphics[width=0.21\linewidth]{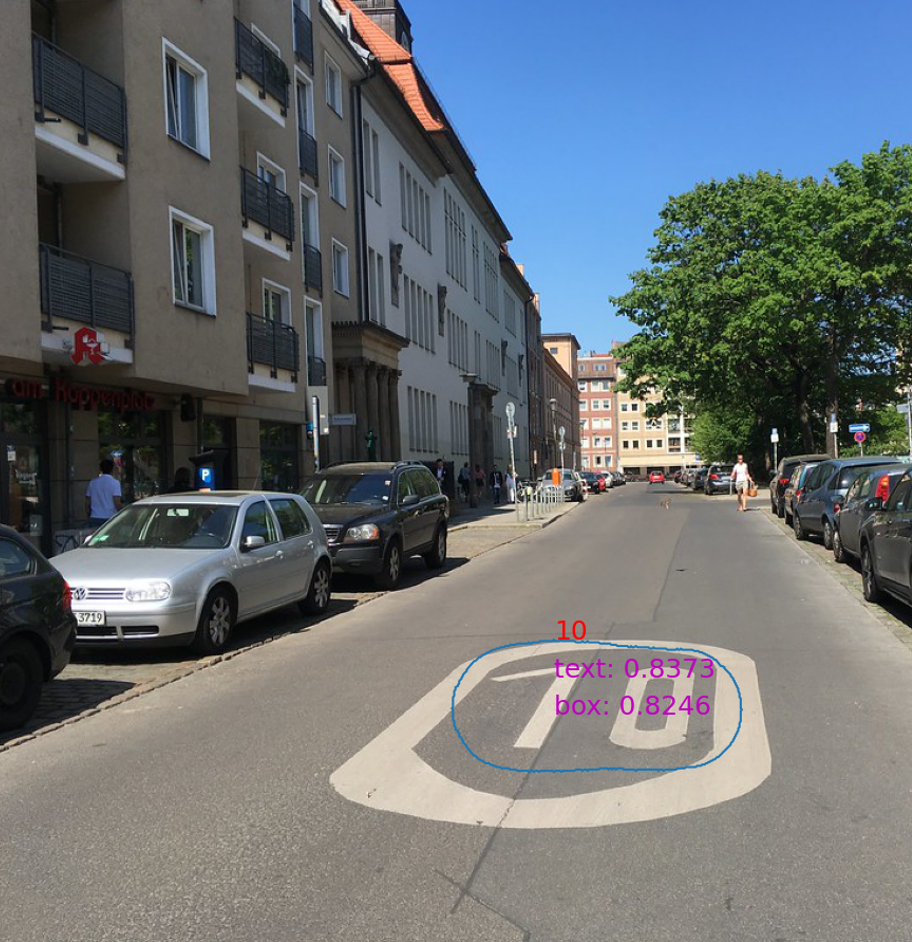}}\hfill
    \caption{Examples of number strings in STR results. The STR results are listed below the corresponding image as well as displayed on the images. }
    \label{fig:num}
\end{figure*}
\begin{figure*}[!]
    \centering    
    \subfigure[]{\includegraphics[width=0.33\linewidth]{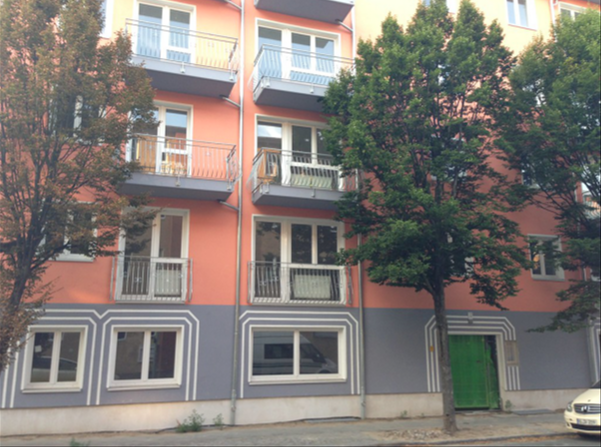}}\hfill
    \subfigure[]{\includegraphics[width=0.33\linewidth]{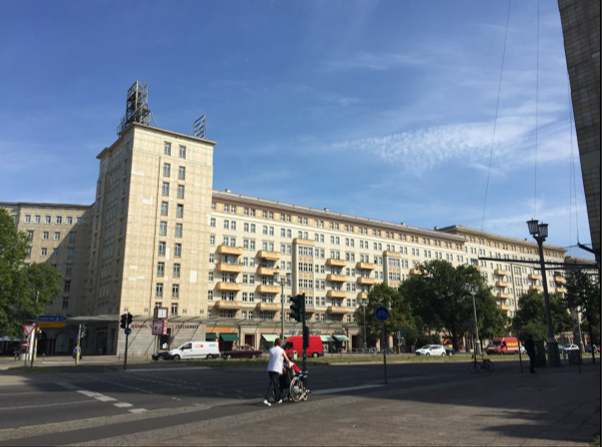}}\hfill
    \subfigure[]{\includegraphics[width=0.33\linewidth]{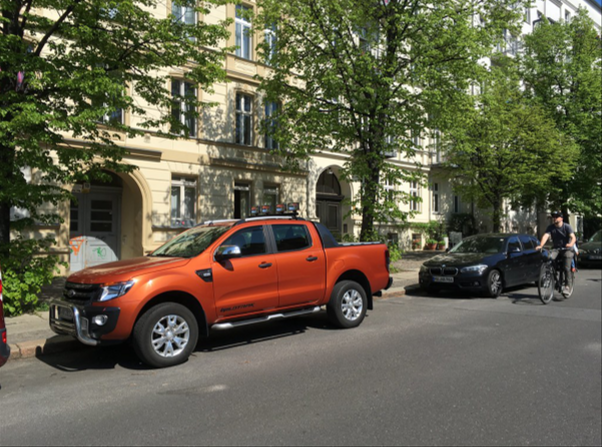}}
    \\
    \subfigure[]{\includegraphics[width=0.33\linewidth]{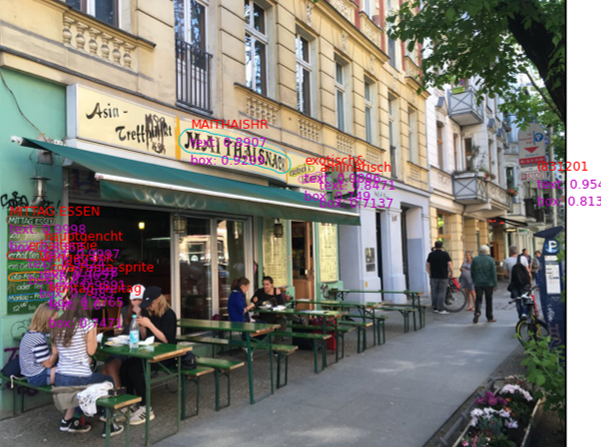}}\hfill
    \subfigure[]{\includegraphics[width=0.33\linewidth]{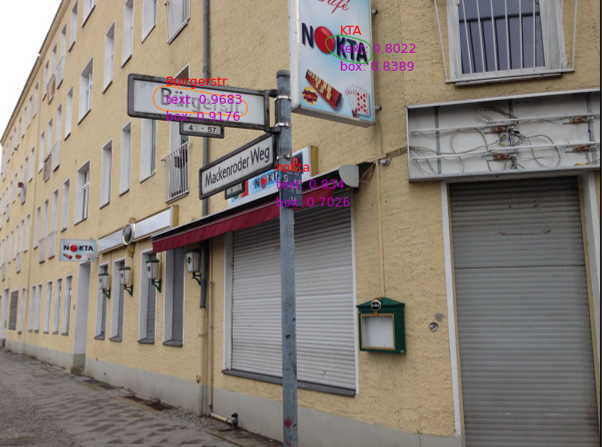}}\hfill
    \subfigure[]{\includegraphics[width=0.33\linewidth]{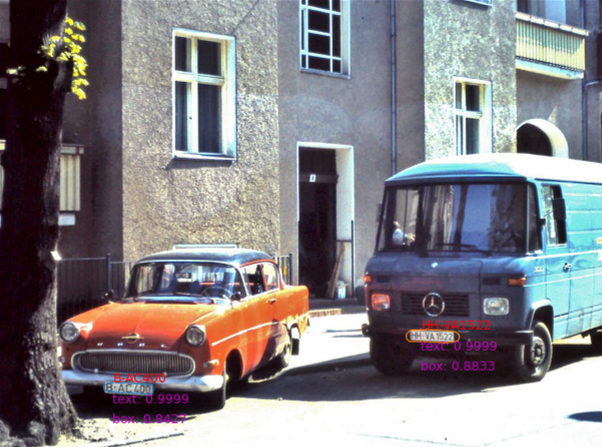}}
    \caption{Examples of residential buildings: (a)(b)(c) have no text recognized in STR results; (d) shows a restaurant on the ground floor of a residential building; (e) and (f) show street name signs and car plates in front of the residential buildings leading to text recognized in STR results. }
    \label{fig:r}
\end{figure*}

\subsection{Experiments and results: STR on Flickr}

First, we applied STR on the Flickr images to extract texts on buildings, using pre-trained models, TextSnake~\citep{long2018textsnake} and SAR~\citep{li2019show} for text detection and recognition, respectively, as introduced in Section~\ref{metho}.  
The results with both text score and box score $>$ 0.8 were kept. 
Second, we filtered out results with stopwords in both English and German, and filtered out text strings with repetitive letters that are not words and were misrecognized from building structures. 

After filtering, valid texts were recognized on 1,558 images (45.4\% of the dataset). The number of recognized texts per image ranges between 1 and 32. 
Figure~\ref{fig:str} shows examples of STR results. 
Table~\ref{tab:overview} summarizes the number of images of the Flickr Berlin dataset and the number of images with STR results for each building type. 
More STR-on-Flickr results can be found at \url{https://github.com/ya0-sun/STR-Berlin}.

\begin{table}[b]
\begin{center}
\begin{tabular}{|c|c|c|} 
 \hline
 %toal & 3,431   & 1,558  \\  
 \textit{building types} & \textit{Flickr images} & \textit{with texts in STR results} \\  
 \hline
 residential    & 1,833     & 892 \\  
 commercial     & 605       & 330  \\  
 other          & 993       & 336 \\ 
 \hline
 total          & 3,431     & 1,558 \\
 \hline
\end{tabular}
\end{center}
\caption{ Number of images of Flickr Berlin dataset and number of images with STR results for each building type.  }
\label{tab:overview}
\end{table}

\subsection{Correctness of numbers recognized by STR}

In 32 images, only numbers were recognized. We manually checked the STR results of these images to verify the results. 

\subsubsection{Correctness of numbers in STR results}

Manual comparison confirmed that STR results on 29 images are correct. 
Although we could not manually label the whole dataset, the correctness of the recognized numbers, i.e., 90.62\%, for the small subset, indicates an overall high accuracy of the STR results.

\subsubsection{Objects containing recognized numbers}

Among the 29 correctly recognized images, 19 are on buildings, i.e., 65.52\%, most of which are house numbers and construction years, e.g., Fig~\ref{fig:num}(a)(b), and a few are on building walls, e.g., Fig~\ref{fig:num}(c). 
Some numbers are on other objects, including static objects, such as speed limit signs and roads (Fig~\ref{fig:num}(d)(e)), and moving objects, such as vehicles and race runners' bibs. Table~\ref{tab:str-num} summarizes the objects on which numbers are recognized. 

The comparison suggests that removing non-building objects in a pre-processing step, e.g., the filtering process in Section~\ref{metho}, is necessary. 

% and on 3 images, STR results are wrong. 

\begin{table}[b]
    \centering
    \begin{tabular}{|l|l|c|}
    \hline
    \multicolumn{2}{|l|}{\textit{Object}}     &   \textit{\#image}\\
    \hline
                    &   house number          &   15  \\
    {\textit{building}}      &   construction year     &   1   \\
                    &   other numbers on walls  &   3   \\
    \hline
    \textit{none-building} &   street furniture      &   2   \\
    \textit{ \& static }   &   road surface marking & 1 \\
    \hline
    \textit{none-building} &   race bib  & 3 \\      
    \textit{\& moving }    &   vehicle  & 2 \\ 
    \hline
    \textit{watermark}     &   timestamp of the photo & 1\\   
    \hline 
    \end{tabular}
    \caption{Objects and the corresponding number of images on which numbers are recognized in STR. }
    \label{tab:str-num}
\end{table}

% !!!texts per certain area!!!
% !!!text size!!!

\subsection{STR results and OSM building functions }

We compared STR results with building function labels from OSM. 

Intuitively, residential buildings are expected to have fewer texts on their facades, and commercial buildings should have more; however, no correlation is observed in our experiments.

\subsubsection{Flickr images with texts recognized by STR} 
STR recognized texts on 892 residential, 330 commercial, and 336 other images, respectively, i.e., about $1/2$ residential and commercial buildings have texts recognized, and the number for other buildings is about $1/3$. 

Residential buildings are expected to have fewer texts. Thus, we investigated Flickr images of residential buildings with STR-recognized texts. Some examples are shown in Figure~\ref{fig:r}. 
We found two main contributors: 
\begin{enumerate}
    \item 
    {Shops on the ground floor or lower floors of residential buildings, as an example given in Figure~\ref{fig:r}(d). The most common businesses include café, restaurant, pharmacy, bakery, butcher, kiosk, and bank. } 
    
    \item 
    {Texts on non-building objects that occlude the buildings. The most common objects are road signs, vehicle registration plates, and reserved parking signs, as examples shown in Figure~\ref{fig:r}(e) and (f). }
\end{enumerate} 

The first case calls for a better definition of building types in OSM, e.g., adding mixed or secondary usage. These buildings have other usages besides their primary usage. 
Object detection can help solve the second case by removing non-building objects.

\subsubsection{Flickr images with no text recognized by STR} 

\begin{table*}[!]
    \centering
    \begin{tabular}{|c|c||c|c|c||c|c|}
    \hline
    & & \multicolumn{3}{|c||}{\textit{Does the image contain:}} & \multicolumn{2}{|c|}{\textit{Is the building function recognizable by:} } \\
    \hline
         & \textit{Flickr example} & \textit{texts?} &  \textit{non-text signals?} & \textit{target building occluded?} & \textit{STR?} & \textit{human?}\\
     \hline
     1  & Figure~\ref{fig:c-no} (a)(b) &  yes & - & - & no & yes\\
     % \hline
     2  & Figure~\ref{fig:c-no} (c)(d)(e)(f)& no & yes & - & no & yes\\
     % \hline
     3  & Figure~\ref{fig:c-no} (g)(h)& no & no & - &  no & no\\
     % \hline
     4  & Figure~\ref{fig:c-no} (i)& - & - & yes &  no & no \\
     \hline
    \end{tabular}
    \caption{Four cases without STR results on Flickr images featuring commercial buildings. }
    \label{tab:c-no}
\end{table*}

Texts were not recognized on $1/2$ residential and commercial building images and $2/3$ other building images.   

\begin{figure*}[]
    \centering    
    \subfigure[]{\includegraphics[width=0.33\linewidth]{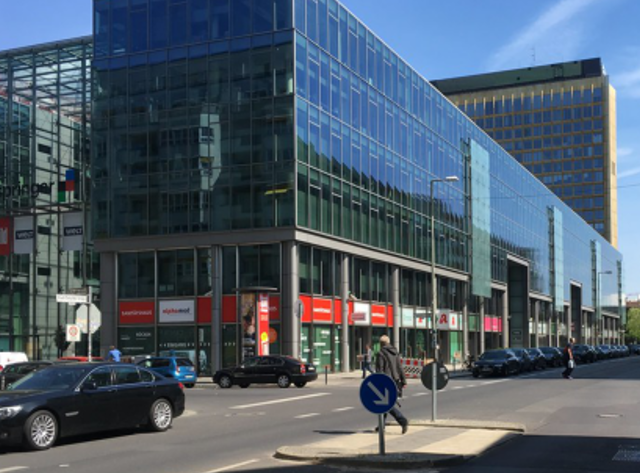}}\hfill
    \subfigure[]{\includegraphics[width=0.33\linewidth]{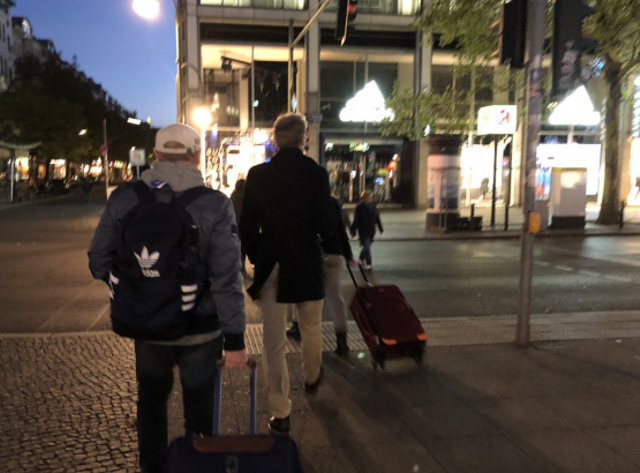}}\hfill
    \subfigure[]{\includegraphics[width=0.33\linewidth]{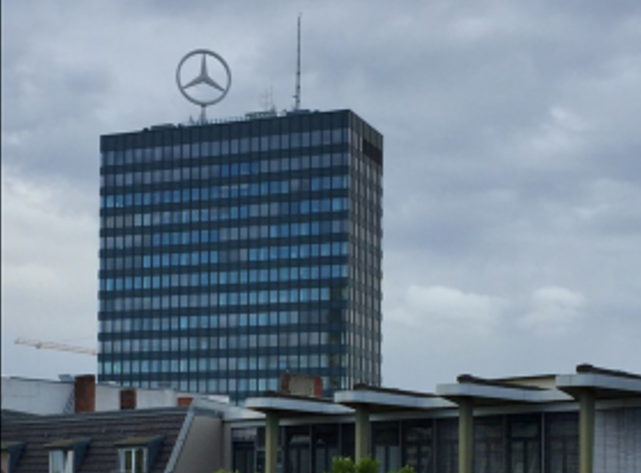}}
    \\
    \subfigure[]{\includegraphics[width=0.33\linewidth]{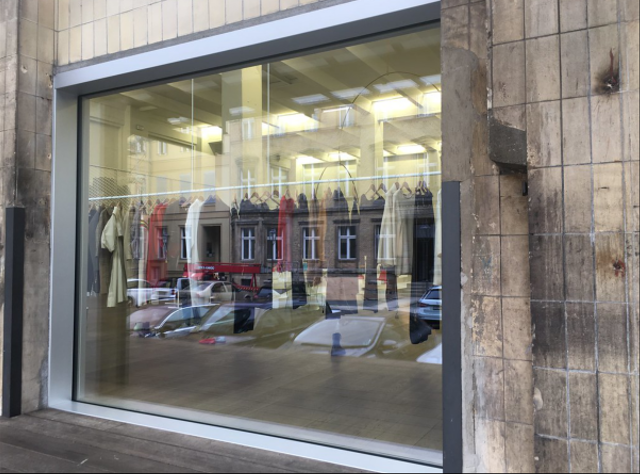}}\hfill
    \subfigure[]{\includegraphics[width=0.33\linewidth]{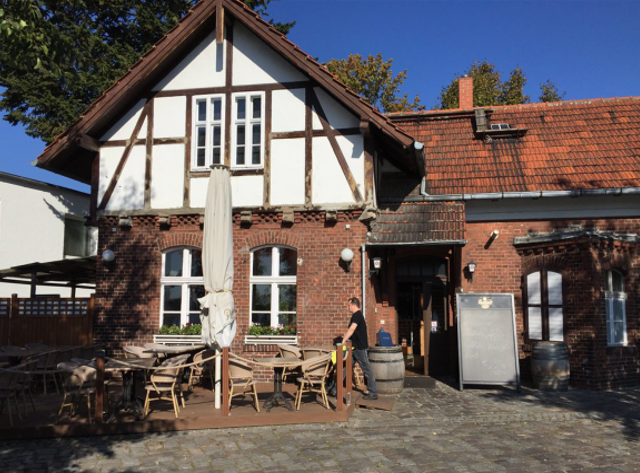}}\hfill
    \subfigure[]{\includegraphics[width=0.33\linewidth]{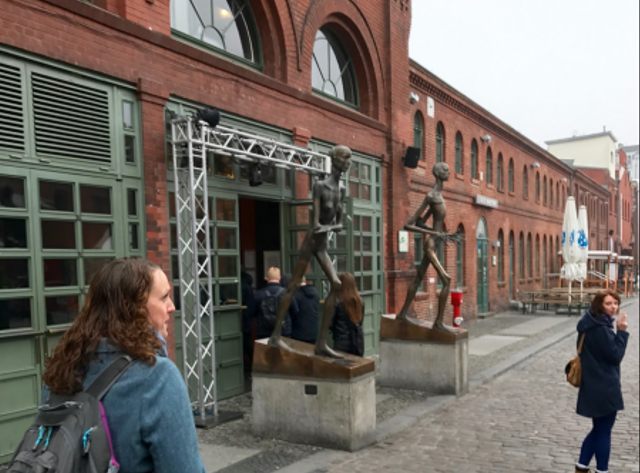}}
    \\
    \subfigure[]{\includegraphics[width=0.33\linewidth]{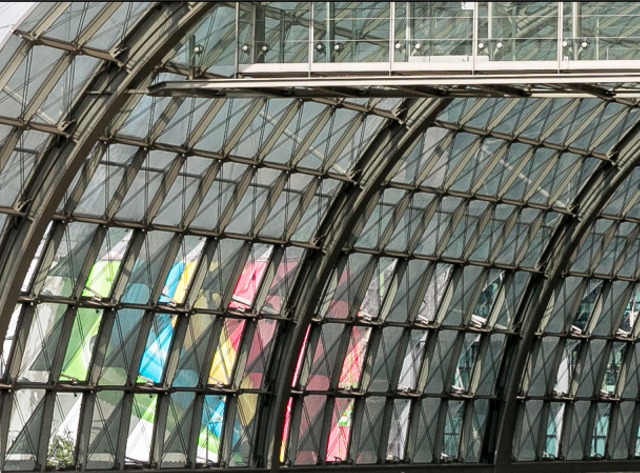}}\hfill
    \subfigure[]{\includegraphics[width=0.33\linewidth]{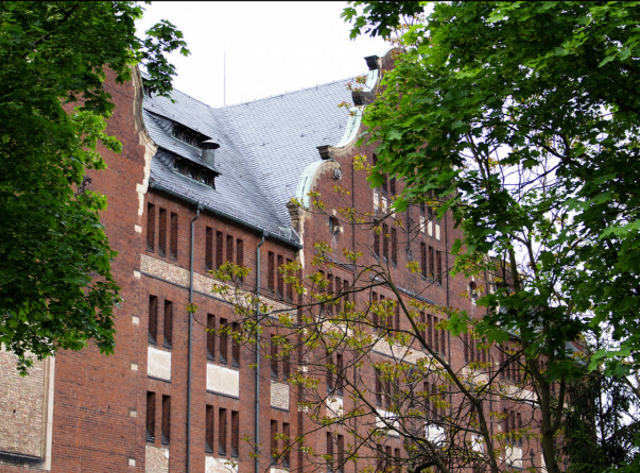}}\hfill
    \subfigure[]{\includegraphics[width=0.33\linewidth]{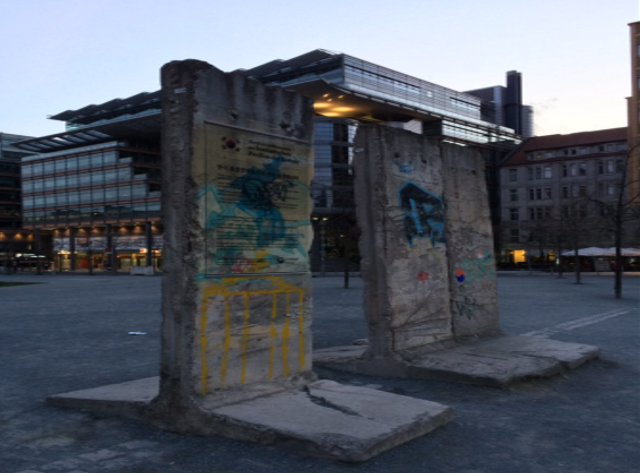}}
    \caption{Examples of commercial buildings with no STR results. }
    \label{fig:c-no}
\end{figure*}

Commercial buildings are expected to have more text on their facades. Thus we further investigated Flickr images of commercial buildings with no text recognized by STR. Some examples are shown in Figure~\ref{fig:c-no}. 
Further investigation of the dataset led us to four cases, listed below and summarized in Table~\ref{tab:c-no}: 

\begin{enumerate}
    \item
    Image contains texts that STR does not recognize because of insufficient image quality, e.g., resolution, light conditions, and blurs, but are easy for humans to recognize building functions. Examples are shown in Figure~\ref{fig:c-no}(a) and (b). 
    
    \item 
    Image contains no text, but other human-readable signals, such as brand symbols, display windows, outdoor restaurant tables, and outside shop statues, can help classify building functions. Some examples shown in Figure~\ref{fig:c-no}(c)(d)(e)(f). 

    \item 
    Image contains neither texts nor other signals, as shown in 
    Figure~\ref{fig:c-no}(g)(h), and it is challenging for humans to classify building functions. 

    \item 
    Buildings in the image are occluded or partially occluded, as shown in 
    Figure~\ref{fig:c-no}(i), and it is difficult for humans to tell building functions. 

\end{enumerate} 

In all the above situations, STR on Flickr, or STR on SVI in more general cases, is unsuitable for building function classification.  
The task's difficulty is especially addressed in Cases 3 and 4, where discerning building types is difficult for humans.

 % table also in fig-c.tex

\subsection{STR failure case analysis}

We observed seven common failure cases in the results that constitute a common challenge for STR on street-view images, and we grouped them below according to the algorithm, the data, and the task, and show some examples in Figure~\ref{fig:w}:

\begin{figure*}[h]
    \centering    
    \subfigure[`Vollkornbackerei', `Steinmuhle']{\includegraphics[width=0.33\linewidth]{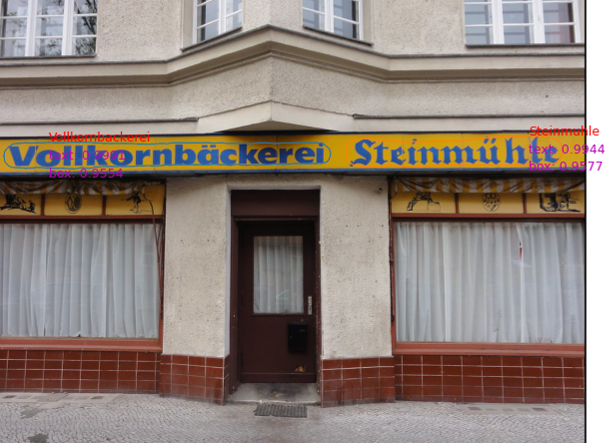}}\hfill
    \subfigure[`record', `store', `CAFE', `BAR', ...]{\includegraphics[width=0.33\linewidth]{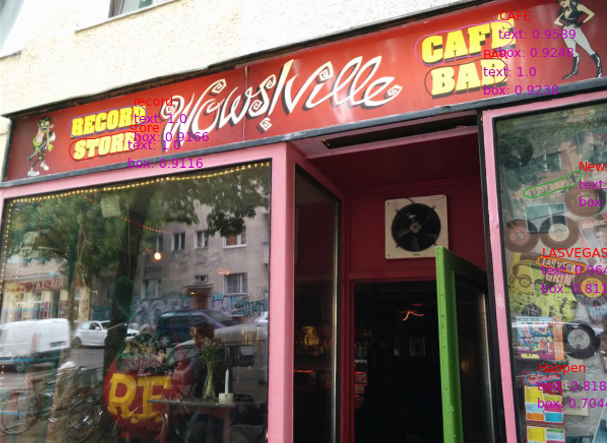}}\hfill
    \subfigure[`hopopol', `BERLINER']{\includegraphics[width=0.33\linewidth]{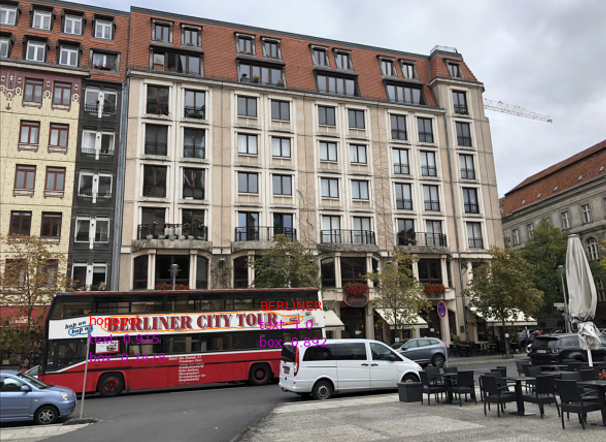}}
    \\
    \subfigure[`Message-ID:']{\includegraphics[width=0.455\linewidth]{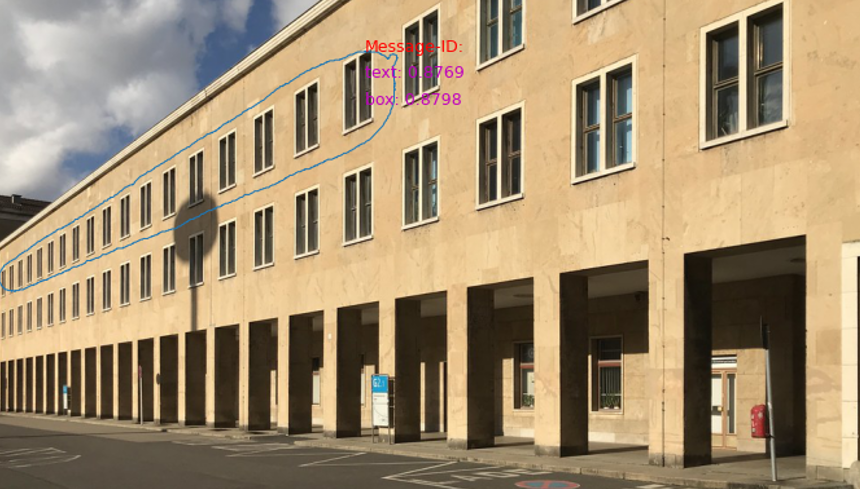}}\hfill
    \subfigure[`1993']{\includegraphics[width=0.29\linewidth]{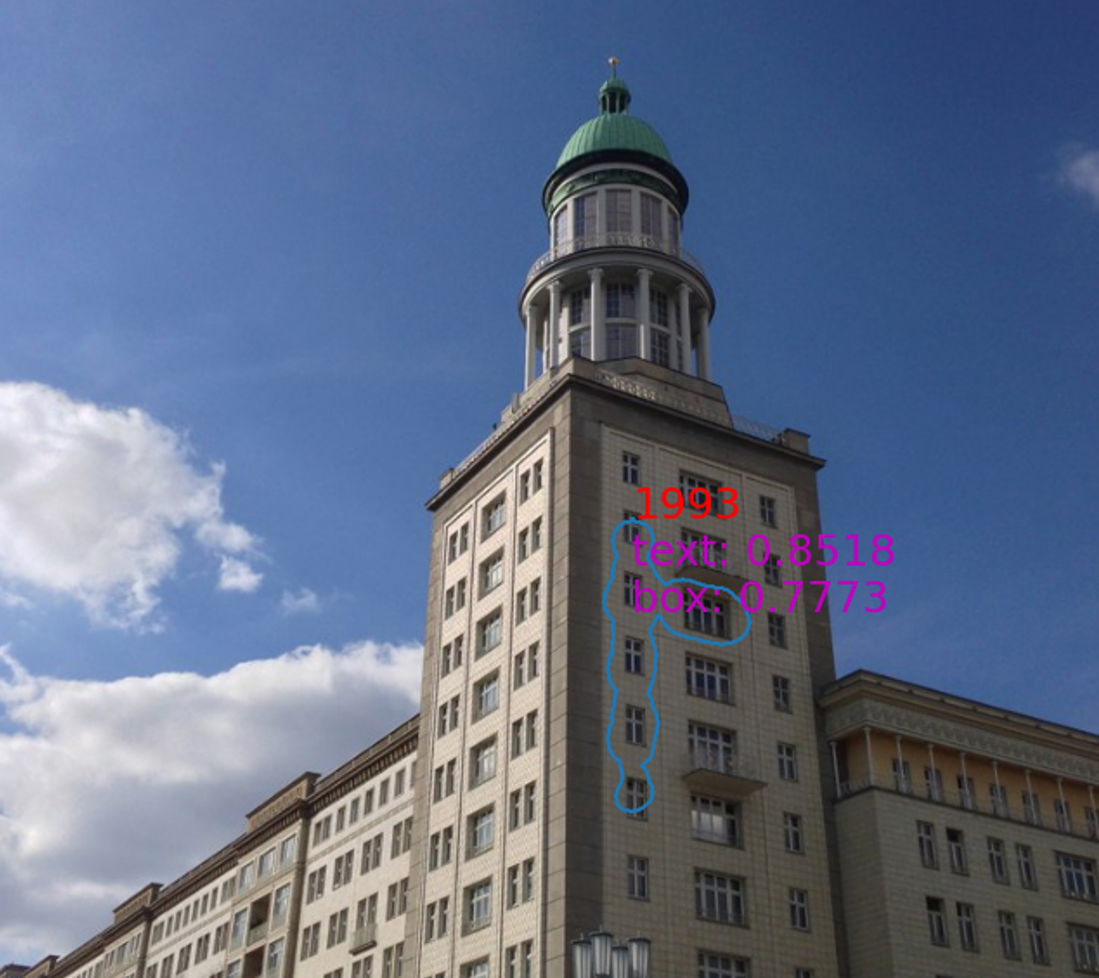}}\hfill
    \subfigure[`HOTEL']{\includegraphics[width=0.25\linewidth]{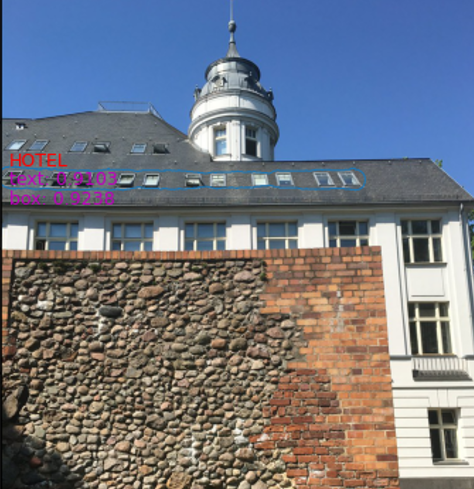}}
    \caption{Examples of failure cases in STR results. The STR results are listed below the corresponding image as well as displayed on the images.}
    \label{fig:w}
\end{figure*}

\begin{enumerate}
\item{Related to STR algorithms}
    \begin{enumerate}
         \item{\textit{Special characters:}} existing models are mostly trained on datasets of English alphanumeric characters, which fail to predict special characters. 
         Figure~\ref{fig:w}(a) shows an example that on the shop sign, `ä' and `ü' in German are recognized as `a' and `u.' 

        \item{\textit{Font styles}}: texts and graffitis are often misrecognized because of font styles. The diverse character expression requires the models to recognize generalized visual features. Such as an example is shown in Figure~\ref{fig:w}(b): texts in regular fonts are recognized, but not the store name between `record' `store' and `CAFE' `BAR.'  

        \item{\textit{Low resolution or lighting condition:}} The models can not handle low-resolution images or photos taken in insufficient lighting conditions. In our test, we accessed the high-resolution Flickr data, but it remains problematic if one uses lower-resolution crowdsourced images, such as Mapillary. Super-resolution modules may improve performance. 
    \end{enumerate}

\item{Related to the images}
    \begin{enumerate}
        \item{\textit{Occlusion:}} partially occluded texts on buildings are not well recognized. Other objects occluding buildings lead to wrong results for building information extraction. Figure~\ref{fig:w}(c) shows an example. 

        \item{\textit{Multiple buildings in one image}}: in the current workflow, one Flickr image is associated with one building footprint in OSM data. However, it is common for multiple buildings to be contained in one image, such that errors occur as texts on one building are mapped in another. Future work should improve the matching between Flickr images and OSM data to solve this problem. 
    
        \item{\textit{Watermark:}} Watermarks on images added by users before uploading or time stamps added by digital cameras often consist of texts or numbers that need to be removed in the STR results. 
                
    \end{enumerate}

\item{Specifically related to the task of STR on SVI}
    \begin{enumerate}
        \textit{Non-text repetitive patterns:}
        we observed that the models misrecognize repetitive patterns on buildings, such as windows and balcony railings, as texts, as shown in Figure~\ref{fig:w}(d)(e)(f). 
        
        The STR-recognized text `HOTEL' in Figure~\ref{fig:w}(f) is particularly misleading and challenging to identify. 
        We suspect that the domain gap between the STR datasets used for training and the street-view images obtained from Flickr be the reason for the observed issues. 
    \end{enumerate}

\end{enumerate}

\section{Conclusion and Future Work}\label{co}

This work explores building attribute mapping with crowdsourced street-view images, specifically focusing on texts on building facades. We create a Berlin Flickr dataset and employ pre-trained STR models for text detection and recognition. Since ground truth labels for texts are unavailable, we manually checked a subset of images recognized by STR, indicating high accuracy. We further analyzed the relationship between building functions and text recognition, finding no clear correlation.

We identified three main reasons that impose challenges to the task of building attribute mapping using STR: 
\textit{First, the discrepancy between street-view images and popular datasets for STR tasks.} 
Images in STR datasets, e.g., COCO Text~\citep{veit2016coco}, are often text-centric, but street-view images often feature more objects in larger areas and have much smaller text regions, making text recognition challenging, especially in complex and cluttered scenes. 
\textit{Second, lack of ground truth labels.} 
Without ground truth, evaluating STR methods on numerous images is unrealistic. Although manual checking on a subset of images with numerical texts shows promising results, further assessment is required to validate performance and subsequently map texts as building attributes. 
\textit{Third, inaccurate mapping between Flickr images and building footprints.} Currently, each Flickr image is associated with only one building footprint, causing a mismatch of texts from other buildings or objects in the image to the building footprint.
Future improvements are needed for matching between Flickr images and OSM building footprints. 
These findings underline the constraints of STR algorithms on crowdsourced street-view images and the importance of labels in scenes, which will be addressed in future works toward large-scale building attribute mapping.

It is worth noting that the STR approach is suited for buildings with visible texts on their facades but not other cases. Considering alternative approaches or data sources is necessary for large area mapping. 
In addition, this study views Flickr data as a valuable resource to expand street view data through crowd-sourcing and did not investigate its distribution within the city. However, considering the relationship of Flickr image locations with hotspots in cities, interdisciplinary collaboration may be essential to understand the reasons behind building photography and labeling in the dataset and the needed approaches to map building attributes not only for hot spots but for the entire city.

\section*{Acknowledgement}
The work is jointly supported by the German Research Foundation (DFG) under the grant ZH 498/14-1 and ME 1846/16-1 for the project OpenStreetMap Boosting using Simulation-Based Remote Sensing Data Fusion (Acronym: OSMSim)
and by the Technical University of Munich (TUM) Georg Nemetschek Institute under the project Artificial Intelligence for the automated creation of multi-scale digital twins of the built world (Acronym:AI4TWINNING). 

\bibliography{ISPRSguidelines_authors}
\end{document}